\documentclass[10pt,twocolumn,letterpaper]{article}

\usepackage{cvpr}
\usepackage{times}
\usepackage{epsfig}
\usepackage{graphicx}
\usepackage{amsmath}
\usepackage{amssymb}
\usepackage{authblk}

\usepackage[colorlinks = true,
            linkcolor = red,
            urlcolor  = blue,
            citecolor = green,
            anchorcolor = blue]{hyperref}

\cvprfinalcopy % *** Uncomment this line for the final submission

\def\cvprPaperID{****} % *** Enter the CVPR Paper ID here

\begin{document}

%%%%%%%%% TITLE
\title{The ActivityNet Large-Scale Activity Recognition Challenge 2018 Summary}

\author[1]{Bernard Ghanem}
\author[2,3]{Juan Carlos Niebles}
\author[4]{Cees Snoek}
\author[1]{Fabian Caba Heilbron}
\author[1]{Humam Alwassel}
\author[1]{Victor Escorcia}
\author[2]{Ranjay Krishna}
\author[2]{Shyamal Buch}
\author[1]{Cuong Duc Dao}

\affil[1]{King Abdullah University of Science and Technology}
\affil[2]{Stanford University}
\affil[3]{Universidad del Norte}
\affil[4]{Universiteit van Amsterdam}

\maketitle
%\thispagestyle{empty}

%%%%%%%%% ABSTRACT
\begin{abstract}
    The 3rd annual installment of the ActivityNet Large-Scale Activity Recognition Challenge, held as a full-day workshop in CVPR 2018, focused on the recognition of daily life, high-level, goal-oriented activities from user-generated videos as those found in internet video portals. The 2018 challenge hosted six diverse tasks which aimed to push the limits of semantic visual understanding of videos as well as bridge visual content with human captions. Three out of the six tasks were based on the ActivityNet dataset, which was introduced in CVPR 2015 and organized hierarchically in a semantic taxonomy. These tasks focused on tracing evidence of activities in time in the form of proposals, class labels, and captions. In this installment of the challenge, we hosted three guest tasks to enrich the understanding of visual information in videos. The guest tasks focused on complementary aspects of the activity recognition problem at large scale and involved three challenging and recently compiled datasets: the Kinetics-600 dataset from Google DeepMind, the AVA dataset from Berkeley and Google, and the Moments in Time dataset from MIT and IBM Research.
\end{abstract}

%%%%%%%%% BODY TEXT
\section{Introduction}
This challenge was the 3rd annual installment of the ActivityNet Large-Scale Activity Recognition Challenge held as a full-day workshop in CVPR 2018. It focused on the recognition of daily life, high-level, goal-oriented activities from user-generated videos as those found in internet video portals. The 2018 challenge hosted six diverse tasks which aimed to push the limits of semantic visual understanding of videos as well as bridge visual content with human captions. Three out of the six tasks were based on the ActivityNet dataset \cite{activitynet}, which was introduced in CVPR 2015 and organized hierarchically in a semantic taxonomy. These tasks focused on tracing evidence of activities in time in the form of proposals, class labels, and captions \cite{krishna2017dense}. In this installment of the challenge, we hosted three guest tasks to enrich the understanding of visual information in videos. The guest tasks focused on complementary aspects of the activity recognition problem at large scale and involved three challenging and recently compiled datasets: the Kinetics-600 dataset \cite{kinetics} from AVA dataset \cite{ava} from Berkeley and Google, and the Moments in Time dataset \cite{momentsintime} from MIT and IBM Research.

\section{Main Challenge Tasks}
The challenge had three main tasks: \textbf{Temporal Action Proposals} (ActivityNet), \textbf{Temporal Action Localization} (ActivityNet), and \textbf{Dense-Captioning Events in Videos} (ActivityNet Captions). In the following subsections, we describe each task's objective, dataset, and evaluation metric. We finally give the top-$3$ results on each task.

\subsection{Task 1: Temporal Action Proposals}
\noindent\textbf{Description and Objective.} In many large-scale video analysis scenarios, one is interested in localizing and recognizing human activities occurring in short temporal intervals within long untrimmed videos. Current approaches for activity detection still struggle to handle large-scale video collections and efficiently addressing this task remains elusive to our visual systems. This is in part due to the computational complexity of current action recognition approaches and the lack of methods that propose fewer intervals in the video, where activity processing can be focused. These set of candidate temporal segments are widely known as \textit{Action Proposals}.

To be applicable at large-scales and in practical scenarios, a useful action proposal method is driven by two competing goals. (i) The proposal method must be computationally efficient in representing, encoding, and scoring a temporal segment. (ii) The proposal method must be discriminative of activities that we are interested in, so as to only retrieve temporal segments that contain visual information indicative of these activity classes. Thus, this task is intended to push the state-of-the-art in action proposal generation algorithms forward.

\noindent\textbf{Dataset.} This task is evaluated on the ActivityNet version 1.3 dataset \cite{activitynet}. The dataset consists of more than $648$ hours of untrimmed videos from a total of ~$20$K videos. It contains $200$ different daily activities such as: \textit{walking the dog}, \textit{long jump}, and \textit{vacuuming floor}. The distribution among training, validation, and testing is roughly $50\%$, $25\%$, and $25\%$ of the total videos, respectively.

\noindent\textbf{Evaluation Metric.} We use the area under the Average Recall vs. Average Number of Proposals per Video (AR-AN) curve as the evaluation metric for this task. A proposal is a true positive if it has a temporal intersection over union (tIoU) with a ground-truth segment that is greater than or equal to a given threshold (\eg tIoU $> 0.5$). AR is defined as the mean of all recall values using tIoU between $0.5$ and $0.95$ (inclusive) with a step size of $0.05$. AN is defined as the total number of proposals divided by the number of videos in the testing subset. We consider $100$ bins for AN, centered at values between $1$ and $100$ (inclusive) with a step size of $1$, when computing the values on the AR-AN curve. 

\noindent\textbf{Top Results.} 
Table \ref{table:task1} shows the top-3 submissions. Please refer to \href{http://activity-net.org/challenges/2018/}{challenge website} for the \textit{full version} of this summary document which includes a copy of all the papers submitted to the workshop.

\begin{table}[h!]
\centering
\begin{tabular}{c c c} 
 \hline
 Rank & Organization & AUC \\ 
 \hline
 1 & Baidu Vis & 71.00 \\ 
 2 & Shanghai Jiao Tong University & 69.30 \\
 3 & YH Technologies & 67.78 \\
 \hline
\end{tabular}
\vspace{3pt}
\caption{The top-3 submissions for task 1.}
\label{table:task1}
\end{table}

% #########
\subsection{Task 2: Temporal Action Localization}
\noindent\textbf{Description and Objective.} Despite the recent advances in large-scale video analysis, temporal action localization remains as one of the most challenging unsolved problems in computer vision. This search problem hinders various real-world applications ranging from consumer video summarization to surveillance, crowd monitoring, and elderly care. Therefore, we are committed to push forward the development of efficient and accurate automated methods that can search and retrieve events and activities in video collections. This task is intended to encourage computer vision researchers to design high performance action localization systems.

\noindent\textbf{Dataset.} This task is evaluated on the ActivityNet version 1.3 dataset \cite{activitynet}. The dataset consists of more than $648$ hours of untrimmed videos from a total of ~$20$K videos. It contains $200$ different daily activities such as: \textit{walking the dog}, \textit{long jump}, and \textit{vacuuming floor}. The distribution among training, validation, and testing is roughly $50\%$, $25\%$, and $25\%$ of the total videos, respectively.

\noindent\textbf{Evaluation Metric.} We use the Interpolated Average Precision (AP) to evaluate the results on each activity category. The performance on the dataset is measured by the mean AP (mAP) over all the activity categories. To determine if a detection is a true positive, we inspect the tIoU with a ground truth segment, and check whether it is greater or equal to a given threshold (\eg tIoU $>0.5$). The official metric used in this task is the average mAP, which is defined as the mean of all mAP values computed with tIoU thresholds between $0.5$ and $0.95$ (inclusive) with a step size of $0.05$.

\noindent\textbf{Top Results.} 
Table \ref{table:task2} shows the top-3 submissions. Please refer to \href{http://activity-net.org/challenges/2018/}{challenge website} for the \textit{full version} of this summary document which includes a copy of all the papers submitted to the workshop.

\begin{table}[h!]
\centering
\begin{tabular}{c c c} 
 \hline
 Rank & Organization & Average mAP \\ 
 \hline
 1 & Shanghai Jiao Tong University & 38.53 \\ 
 2 & YH Technologies & 35.49 \\
 3 & Baidu Vis & 35.27 \\
 \hline
\end{tabular}
\vspace{3pt}
\caption{The top-3 submissions for task 2.}
\label{table:task2}
\end{table}

% ##################
\subsection{Task 3: Dense-Captioning Events in Videos}
\noindent\textbf{Description and Objective.}  Most natural videos contain numerous events. For example, in a video of a \textit{man playing a piano}, the video might also contain another \textit{man dancing} or a \textit{crowd clapping}. This task aims to tackle the challenges of dense-captioning events, which involves both detecting and describing events in a video. 

\noindent\textbf{Dataset.} This task is evaluated on the ActivityNet Captions dataset \cite{krishna2017dense}. The dataset connects videos to a series of temporally annotated sentence descriptions. Each sentence covers a unique segment of the video, describing multiple events that occur. These events may occur over very long or short periods of time and are not limited in any capacity, allowing them to co-occur. On average, each of the $20$K videos in ActivityNet Captions contains $3.65$ temporally localized sentences, resulting in a total of $100$K sentences. The number of sentences per video follows a relatively normal distribution. Furthermore, as the video duration increases, the number of sentences also increases. Each sentence has an average length of $13.48$ words, which is also normally distributed.

\noindent\textbf{Evaluation Metric.} Inspired by the dense-image captioning metric, we use a similar metric to measure the joint ability of our model to both localize and caption events. This metric computes the average precision (AP) across tIoU thresholds of $0.3$, $0.5$, and $0.7$, when captioning the top-$1000$ proposals. We measure precision of captions using the traditional evaluation metrics: Bleu, METEOR and CIDEr.

\noindent\textbf{Top Results.} 
Table \ref{table:task3} shows the top-2 submissions. Please refer to \href{http://activity-net.org/challenges/2018/}{challenge website} for the \textit{full version} of this summary document which includes a copy of all the papers submitted to the workshop.

\begin{table}[h!]
\centering
\begin{tabular}{c c c} 
 \hline
 Rank & Organization & Average Meteor \\ 
 \hline
 1 & RUC and CMU & 8.53 \\ 
 2 & YH Technologies & 8.13 \\
 \hline
\end{tabular}
\vspace{3pt}
\caption{The top-2 submissions for task 3.}
\label{table:task3}
\end{table}

% ##################
\section{Hosted Challenge Tasks}
In this installment of the challenge, we hosted three guest tasks to enrich the understanding of visual information in videos. These guest tasks focused on complementary aspects of the activity recognition problem at large scale and involved three challenging and recently compiled datasets: the Kinetics-600 dataset \cite{kinetics} from Google DeepMind, the AVA dataset \cite{ava} from Berkeley and Google, and the Moments in Time dataset \cite{momentsintime} from MIT and IBM Research.

\subsection{Task A: Trimmed Activity Recognition}
\noindent\textbf{Description and Objective.} This task is intended to evaluate the ability of algorithms to recognize activities in trimmed video sequences. Here, videos contain a single activity, and all the clips have a standard duration.

\noindent\textbf{Dataset.} This task is evaluated on the Kinetics-600 dataset \cite{kinetics}. Kinetics is a large-scale, high-quality dataset of YouTube video URLs which include a diverse range of human focused actions. The dataset consists of approximately $500$K video clips, and covers $600$ human action classes with at least $600$ video clips for each action class. Each clip lasts around 10s and is labeled with a single class. All of the clips have been through multiple rounds of human annotation, and each is taken from a unique YouTube video. The actions cover a broad range of classes including human-object interactions such as \textit{playing instruments}, as well as human-human interactions such as \textit{shaking hands} and \textit{hugging}.

\noindent\textbf{Evaluation Metric.} We use the top-$k$ accuracy on the testing set as the official metrics for this task. For each video, an algorithm should produce $k$ labels $l_{j}$, $j = 1,..,k$. The ground truth label for the video is $g$. The error of the algorithm for that video would be: $e = \min_{j} d(l_{j}, g),$ with $d(x,y) = 0$ if $x=y$ and $1$ otherwise. The overall error score for an algorithm is the average error over all videos. We will use $k=1$ and $k=5$ and the winner of the challenge is selected based on the average of these two errors.

\noindent\textbf{Top Results.} 
Table \ref{table:taskA} shows the top-3 submissions. Please refer to \href{http://activity-net.org/challenges/2018/}{challenge website} for the \textit{full version} of this summary document which includes a copy of all the papers submitted to the workshop.

\begin{table}[h!]
\centering
\begin{tabular}{c c c} 
 \hline
 Rank & Organization & Average Error \\ 
 \hline
 1 & Baidu Vis & 10.99 \\ 
 2 & YH Technologies & 11.69 \\
 3 & QINIU and SARI & 12.20 \\
 \hline
\end{tabular}
\vspace{3pt}
\caption{The top-3 submissions for task A.}
\label{table:taskA}
\end{table}

% ##############
\subsection{Task B: Spatio-temporal Action Localization}
\noindent\textbf{Description and Objective.} This task is intended to evaluate the ability of algorithms to localize human actions in space and time. Each labeled video segment can contain multiple subjects, each performing potentially multiple actions. The goal is to identify these subjects and actions over continuous video clips. This task is divided into two tracks. Track \#1 is strictly computer vision, \ie participants are requested not to use signals derived from audio, metadata, etc. Track \#2 lifts this restriction, allowing creative solutions that leverage any input modalities.

\noindent\textbf{Dataset.} This task is evaluated on the AVA Dataset version v2.1 \cite{ava}. The AVA dataset densely annotates $80$ atomic visual actions in $430$ $15$-minute movie clips, where actions are localized in space and time, resulting in $1.58$M action labels with multiple labels per human occurring frequently. Clips are drawn from contiguous segments of movies, to open the door for temporal reasoning about activities. The dataset is split into $235$ videos for training, $64$ videos for validation, and $131$ videos for test.

\noindent\textbf{Evaluation Metric.} We use the Frame-mAP at spatial IoU $\ge 0.5$ as the metric for evaluating algorithms on this task. Since action frequency in AVA follows the natural distribution, the metric is averaged across the top $60$ most common action classes in AVA.

\noindent\textbf{Top Results.} 
Tables \ref{table:taskB1} and \ref{table:taskB2} show the top-3 submissions for each track. Please refer to \href{http://activity-net.org/challenges/2018/}{challenge website} for the \textit{full version} of this summary document which includes a copy of all the papers submitted to the workshop.

\begin{table}[h!]
\centering
\begin{tabular}{c c c} 
 \hline
 Rank & Organization & mAP@0.5IoU  \\ 
 \hline
 1 & Tsinghua University & 21.08 \\ 
 2 & Google DeepMind & 21.03 \\
 3 & YH Technologies & 19.60 \\
 \hline
\end{tabular}
\vspace{3pt}
\caption{The top-3 submissions for task B (computer vision only track).}
\label{table:taskB1}
\end{table}

\begin{table}[h!]
\centering
\begin{tabular}{c c c} 
 \hline
 Rank & Organization & mAP@0.5IoU  \\ 
 \hline
 1 & Tsinghua University & 20.99 \\ 
 2 & YH Technologies & 19.60 \\
 3 & UMD & 16.76 \\
 \hline
\end{tabular}
\vspace{3pt}
\caption{The top-3 submissions for task B (full track).}
\label{table:taskB2}
\end{table}

% ##########################
\subsection{Task C: Trimmed Event Recognition}
\noindent\textbf{Description and Objective.} This task is intended to evaluate the ability of algorithms to classify events in trimmed $3$-second videos. Here, videos contain a single activity, and all clips have a standard duration of $3$ seconds. This task is divided into two tracks. The first track uses the Moments in Time dataset \cite{momentsintime}, a new large-scale dataset for video understanding, which has $800$K videos in the training set. The second track use the Moments in Time Mini dataset, a subset of Moments in Time with $100$k videos provided in the training set.

\noindent\textbf{Dataset.} This task is evaluated on the Moments in Time Dataset \cite{momentsintime}. Moments in Time Dataset is a large-scale collection of $1$M $3$-second videos corresponding to spatial-audio-temporal events.

\noindent\textbf{Evaluation Metric.} We use the top-$k$ accuracy on the testing set as the official metrics for this task. For each video, an algorithm should produce $k$ labels $l_{j}$, $j = 1,..,k$. The ground truth label for the video is $g$. The error of the algorithm for that video would be: $e = \min_{j} d(l_{j}, g),$ with $d(x,y) = 0$ if $x=y$ and $1$ otherwise. The overall error score for an algorithm is the average error over all videos. We will use $k=1$ and $k=5$ and the winner of the challenge is selected based on the average of these two errors.

\noindent\textbf{Top Results.} 
Tables \ref{table:taskC1} and \ref{table:taskC2} show the top-3 submissions for each track. Please refer to \href{http://activity-net.org/challenges/2018/}{challenge website} for the \textit{full version} of this summary document which includes a copy of all the papers submitted to the workshop.
\begin{table}[h!]
\centering
\begin{tabular}{c c c} 
 \hline
 Rank & Organization & Average Accuracy  \\ 
 \hline
 1 & Hikvision & 52.91 \\ 
 2 & Megvii & 51.26 \\ 
 3 & Qiniu AtLab & 50.06 \\ 
 \hline
\end{tabular}
\vspace{3pt}
\caption{The top-3 submissions for task C (full track).}
\label{table:taskC1}
\end{table}

\begin{table}[h!]
\centering
\begin{tabular}{c c c} 
 \hline
 Rank & Organization & Average Accuracy  \\ 
 \hline
 1 & Sun Yat-Sen University & 47.72 \\ 
 2 & Beihang University & 45.49 \\ 
 3 & National Taiwan University & 45.10 \\ 
 \hline
\end{tabular}
\vspace{3pt}
\caption{The top-3 submissions for task C (mini track).}
\label{table:taskC2}
\end{table}

{\small
\bibliographystyle{ieee}
\bibliography{egbib}
}

\end{document}